\documentclass{article}
\usepackage[final]{neurips_2021}
\usepackage{url}
\usepackage{amsfonts}       % blackboard math symbols
\usepackage{amsmath,amssymb}
\usepackage{amsthm}
\usepackage{nicefrac}       % compact symbols for 1/2, etc.
\usepackage{microtype}      % microtypography
\usepackage{url}
\usepackage{array,booktabs,makecell}
\usepackage{natbib}
\usepackage{hyperref}
\usepackage[capitalize]{cleveref}
\usepackage{autonum}
\usepackage{mathtools}
\usepackage[colorinlistoftodos]{todonotes}
\usepackage{enumitem}
\usepackage[parfill]{parskip}
\usepackage{bm}
\usepackage{mathrsfs}
\usepackage{subfigure} 
\usepackage{dblfloatfix}
\usepackage{multirow}
\usepackage{wrapfig}
\usepackage{multicol}
\usepackage{caption}

\usepackage[scr=esstix]{mathalfa} %symbols
\usepackage{mathtools}
\usepackage{float}
\graphicspath{{figures/}}

\usepackage{algorithmic}
\usepackage{algorithm}
\usepackage{changepage}
\newtheorem{theorem}{Theorem}

\newtheorem{assumption}{Assumption}

\title{Reinforcement Learning in Factored Action Spaces using Tensor Decompositions}

\author{Anuj Mahajan\thanks{Correspondence to \texttt{ anuj.mahajan@cs.ox.ac.uk}}\hspace{1mm}\thanks{University of Oxford} \And Mikayel Samvelyan \thanks{UCL} \And Lei Mao\thanks{NVIDIA} \And Viktor Makoviychuk\footnotemark[4] \And Animesh Garg\footnotemark[4] \And Jean Kossaifi\footnotemark[4] \And Shimon Whiteson\footnotemark[2] \And Yuke Zhu\footnotemark[4] \And Animashree Anandkumar\footnotemark[4] }

\begin{document}	
\maketitle
% % Sets the space above and below single-line equations.
% \setlength{\abovedisplayskip}{4pt}
% \setlength{\belowdisplayskip}{4pt}

% \title{Tesseract: Tensorised Actors for Multi-Agent Reinforcement Learning}

% \begin{icmlauthorlist}
% \icmlauthor{Anuj Mahajan}{ox}
% \icmlauthor{Mikayel Samvelyan}{}
% \icmlauthor{Lei Mao}{}
% \icmlauthor{Viktor Makoviychuk}{}
% \icmlauthor{Animesh Garg}{}
% \icmlauthor{Jean Kossaifi}{}
% \icmlauthor{Shimon Whiteson}{}
% \icmlauthor{Yuke Zhu}{}
% \icmlauthor{Animashree Anandkumar}{}
% \end{icmlauthorlist}

% \begin{icmlauthorlist}
% \icmlauthor{Anuj Mahajan}{ox}
% \icmlauthor{Mikayel Samvelyan}{uc}
% \icmlauthor{Lei Mao}{nv}
% \icmlauthor{Viktor Makoviychuk}{nv}
% \icmlauthor{Animesh Garg}{nv}
% \icmlauthor{Jean Kossaifi}{nv}
% \icmlauthor{Shimon Whiteson}{ox}
% \icmlauthor{Yuke Zhu}{nv}
% \icmlauthor{Animashree Anandkumar}{nv}
% \end{icmlauthorlist}

% \icmlaffiliation{ox}{University of Oxford}
% \icmlaffiliation{nv}{NVIDIA}
% \icmlaffiliation{uc}{University College London}

% \icmlcorrespondingauthor{Anuj Mahajan}{anuj.mahajan@cs.ox.ac.uk}

\begin{abstract}
We present an extended abstract for the previously published work \textsc{Tesseract}~\citep{pmlr-v139-mahajan21a}, which proposes a novel solution for Reinforcement Learning (RL) in large, factored action spaces using tensor decompositions. The goal of this abstract is twofold: (1) To garner greater interest amongst the tensor research community for creating methods and analysis for approximate RL, (2) To elucidate the generalised setting of factored action spaces where tensor decompositions can be used. 
% beyond decentralised multi-agent RL. 
We use cooperative multi-agent reinforcement learning scenario as the exemplary setting where the action space is naturally factored across agents and learning becomes intractable without resorting to approximation on the underlying hypothesis space for candidate solutions. 
% Reinforcement Learning in large action spaces is a challenging problem. Cooperative multi-agent reinforcement learning (MARL) exacerbates matters by imposing various constraints on communication and observability.In this work, we consider the fundamental hurdle affecting both value-based and policy-gradient approaches: an exponential blowup of the action space with the number of agents. For value-based methods, it poses challenges in accurately representing the optimal value function. For policy gradient methods, it makes training the critic difficult and exacerbates the problem of the \emph{lagging} critic. We show that from a learning theory perspective, both problems can be addressed by accurately representing the associated action-value function with a low-complexity  hypothesis class. This requires accurately modelling the agent interactions in a sample efficient way. To this end, we propose a novel tensorised formulation of the Bellman equation. This gives rise to our method \textsc{Tesseract}, which views the $Q$-function as a tensor whose modes correspond to the action spaces of different agents. Algorithms derived from \textsc{Tesseract} decompose the $Q$-tensor across agents and utilise low-rank tensor approximations to model  agent interactions relevant to the task.  We provide PAC analysis for \textsc{Tesseract}-based algorithms and highlight their relevance to the class of rich observation MDPs. Empirical results in different domains confirm \textsc{Tesseract}'s gains in sample efficiency predicted by the theory.
\end{abstract}
	
\section{Introduction}
\label{intro}
Reinforcement learning (RL) has experienced tremendous advancements in recent years towards coming up with agents which can solve complex tasks and demonstrate generally performant intelligent behaviour \citep{silver2017mastering,vinyals2019grandmaster, oel2021}. One dimension in which task complexity in RL has seen big growth is that of size of the action space over which an agent (or a group of agents) have to make decisions. Thus, a lot of research efforts are being made towards creating approximation methods for learning in large action spaces which would otherwise be unnameable for classical methods. This has set the stage for development of innovative methods, analysis and addressal of new challenges which arise from the above approximation. Factorisation of the action space is one such strategy for overcoming computational intractability and learning approximately optimal policies. Factorisation involves decomposing/partitioning the action space into smaller sets so that RL can be done tractably on these smaller sets. This naturally provides a multi-agent perspective to the problem as each factored component can be seen as an agent working jointly with other such agent towards maximising the expected cumulative rewards. Factorisation can be imposed on an RL problem for tractability or it may arise from the problem/environment constraints such as communication limitations, partial observability and multi-agentness. In this work we will be focussing on product based factorisation 
% \footnote{Note that this subsumes sum decomposition when allowing constraints} 
of the action space $\mathbf{U}$ that can be decomposed as cross product over action sets $\{U_i\}_1^n$ such that $\mathbf{U} = \bigotimes_{i=1}^{n}U_i$. This form of decomposition naturally occurs in many application like robotics where action space is multi-dimensional (ex. a single robot has a joint-policy for locomotion and communication) and swarm robotics\cite{bucsoniu2010multi} where each robot signifies a factor in decomposition. 

We will next see how \textsc{Tesseract}~\citep{pmlr-v139-mahajan21a} utilises the structure present in the above factored action spaces using a novel tensorised form of the Bellman equation. This novel form allows for creating algorithms for RL which can provide exponential gain in sample efficiency in the number of factors under the PAC framework. Further it opens up opportunities for using various forms of tensor decomposition for approximation based on the problem structure.

\section{Background}
\label{background}

\paragraph{Factored action spaces}
We consider the simplest setting for factored action space (FAS) where the state of the system is completely observed and is available for deciding an action from each component in the action space factorisation (we refer to this as an 'agent' from hereon, implying the factored problem has $n$ agents, one for each factor). We refer the reader to \cite{pmlr-v139-mahajan21a} for the more general scenarios. For the above scenario, a FAS can be modelled using a multi-agent Markov Decision Process (MMDP) which is defined as the tuple: $\left\langle S,U,P,r,n,\gamma\right\rangle$. Here, S is the state space of the environment with joint action space $\mathbf{U} = U^n$.   
% Formally we define FAS as a tuple : $\left\langle S,\mathbf{U},P,r,n,\{U_i\}_1^n,\gamma\right\rangle$.
Note that we have assumed without loss of generality that each factor/agent admits the same set of actions $U$ ie. $U_i = U \forall i \in \mathcal{A}\equiv \{1,...,n\}$. A policy in FAS is the mapping $\pi: S \to \mathcal{P}(\mathbf{U})$ which gives a distribution over the joint action space given a state. At each time step $t$, an action $u^i \in U$ is chosen for every agent $ i \in \mathcal{A}$ using $\pi$ which forms the joint action $\mathbf{u}\in\mathbf{U}\equiv U^n$.
% $P(s'|s,\mathbf{u}):S\times\mathbf{U}\times S\rightarrow [0,1]$ is the state transition function. $r(s,\mathbf{u}):S \times \mathbf{U} \rightarrow [0,1]$ is the reward function shared by all agents and $\gamma \in [0,1)$ is the discount factor. 
The joint \textit{action-value function} given a policy $\pi$ is defined as: $Q^\pi(s_t, \mathbf{u}_t)=\mathbb{E}_{s_{t+1:\infty},\mathbf{u}_{t+1:\infty}} \left[\sum^{\infty}_{k=0}\gamma^kr_{t+k}|s_t,\mathbf{u}_t\right]$. In RL, the goal is to find the optimal policy $\pi^{*}$ corresponding to the optimal action value function $Q^*$. 
% For the special learning scenario called Centralised Training with Decentralised Execution (CTDE), the learning algorithm has access to the action-observation histories of all agents and the full state during training phase. However, each agent can only condition on its own local action-observation history $\tau^i$  during the decentralised execution phase.
% \vspace{-1mm}
\paragraph{Reinforcement Learning Methods\label{subsec: rl}} RL methods come in a wide variety, but can fundamentally be classified as model based methods and model free methods. Model based methods typically estimate the underlying dynamics of the MDP ($P, r$) whereas model free methods implicitly account for them.
Another important dimension of variance is the algorithmic setup used for learning which differentiates the methods as policy based, value based or actor-critic (a hybrid of previous two). \cite{sutton2011reinforcement} gives a comprehensive overview of these methods. Value-based and actor-critic methods which can attributed most of the recent progress in RL for large action spaces, both rely on an estimator for the action-value function $Q^\pi$ given a target policy $\pi$. $Q^\pi$ satisfies the (scalar)-Bellman expectation equation:
$ Q^{\pi}(s, \mathbf{u}) = r(s, \mathbf{u})+ \gamma\mathbb{E}_{s',\mathbf{u}'}[Q^{\pi}(s', \mathbf{u}')],$ which can equivalently be written in vectorised form as:
\begin{align}
\label{eq:bell}
Q^{\pi} = R + \gamma P^{\pi}Q^{\pi}, 
\end{align}
where $R$ is the mean reward vector of size $S$, $P^{\pi}$ is the transition matrix. The operation on RHS % of \cref{eq:bell}
$\mathcal{T}^{\pi}(\cdot) \triangleq R + \gamma P^{\pi}(\cdot)$ is the Bellman expectation operator for the policy $\pi$. However, \cref{eq:bell} doesn't expose the structure present in FAS problems, in \cref{method} we discuss how \textsc{Tesseract} generalises the Bellman expectation equation (and analogously the Bellman optimality equation) to a novel tensor form suitable for sample efficient learning in FAS. 

\paragraph{Tensor Decomposition}
\begin{wrapfigure}{r}{0.45\linewidth}
\centering
\vspace{-2mm}
\includegraphics[width=\linewidth]{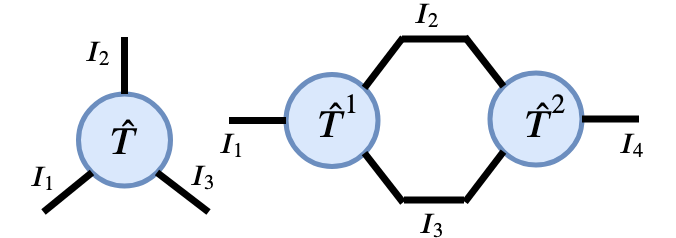}
\caption{Left: Tensor diagram for an order $3$ tensor $\hat T$. Right: Contraction between $\hat T^1$,$\hat T^2$ on common index sets $I_2,I_3$. \label{fig:tdia}}
\vspace{-3mm}
\end{wrapfigure}
Tensors are high dimensional analogues of matrices and tensor methods generalize matrix algebraic operations to higher orders. In the rest of this paper, we use $\hat{\cdot}$ to denote tensors.
Formally, an order $n$ tensor $\hat{T}$ has $n$ index sets ${I}_j,\forall j \in\{1..n\}$ and has elements $T(e), \forall e \in \times_{\mathcal{I}} {I}_j $ taking values in a given set $\mathcal{S}$, where $\times$ is the set cross product and we denote the set of index sets by $\mathcal{I}$. Each dimension $\{1..n\}$ is also called a mode.
% \begin{figure}[h]
% \begin{wrapfigure}{r}{0.4\linewidth}
% \centering
% %\vspace{-2mm}
% \includegraphics[width=\linewidth]{figures/diagram.png}
% \caption{Left: Tensor diagram for an order $3$ tensor $\hat T$. Right: Contraction between $\hat T^1$,$\hat T^2$ on common index sets $I_2,I_3$. \label{fig:tdia}}
%\vspace{-4mm}
% \end{wrapfigure}
% \end{figure}
An elegant way of representing tensors and associated operations is via tensor diagrams as shown in \cref{fig:tdia}. 
Tensor contraction generalizes the concept of matrix with matrix multiplication. 
For any two tensors $\hat{T}^1$ and $\hat{T}^2$ with $\mathcal{I}_{\cap} = \mathcal{I}^1 \cap \mathcal{I}^2$ we define the contraction operation as $\hat T= \hat{T}^1{\odot} \hat{T}^2$ with $ \hat{T}(e_1,e_2) = \sum_{e\in \times_{\mathcal{I}_{\cap}} {I}_j } \hat T^1(e_1,e)\cdot\hat T^2(e_2,e), e_i \in \times_{\mathcal{I}^i\setminus \mathcal{I}_{\cap}} {I}_j$.  
Using this building block, we can define tensor decompositions, which factorizes a (low-rank) tensor in a compact form. This can be done with various decompositions~\citep{kolda2009tensor, janzamin2020spectral}, such as Tucker, Tensor-Train (also known as Matrix-Product-State), or CP (for Canonical-Polyadic). In this paper, we focus on the latter, which we briefly introduce here.
Just as a matrix can be factored as a sum of rank-$1$ matrices (each being an outer product of vectors),  a tensor can be factored as a sum of rank-1 tensors, the latter being an outer product of vectors. The number of terms in the sum is called the \emph{CP-rank}. Formally, a tensor $\hat T$ can be factored using a (rank--$k$) CP decomposition into a sum of $k$ vector outer products (denoted by $\otimes$), as, 
\begin{align}
\label{CPD}
\hat T=\sum_{r=1}^k w_r\otimes^n u_r^i ,i \in \{1..n\},||u_r^i||_2 =1.
\end{align}	
\section{Methodology}
\label{method}
\subsection{Tensorised Bellman equation}

In this section, we provide the basic framework for Tesseract. We focus here on the discrete action space. The extension for continuous actions can be found in \cite{pmlr-v139-mahajan21a}.
% \begin{proposition}
% Any real-valued function $f$ of $n$ arguments $(x_1..x_n)$ each taking values in a finite set $x_i\in \mathcal{D}_i$ can be represented as a tensor $\hat f$ with modes corresponding to the domain sets $\mathcal{D}_i$ and entries $\hat f(x_1..x_n) = f(x_1..x_n)$.  
% \end{proposition}
Given a multi-agent problem $G=\left\langle S,U,P,r,n,\gamma\right\rangle$, let $\mathcal{Q} \triangleq \{Q: S\times U^n\to \mathbb{R}\}$ be the set of real-valued functions on the state-action space. We are interested in the \emph{curried }\cite{barendregt1984introduction} form $Q: S\to U^n\to \mathbb{R},Q\in \mathcal{Q}$ so that $Q(s)$ is an order $n$ tensor (We use functions and tensors interchangeably where it is clear from context). Algorithms in Tesseract operate directly on the curried form and preserve the structure implicit in the output tensor. (Currying in the context of tensors implies fixing the value of some index. Thus, Tesseract-based methods keep action indices free and fix only state-dependent indices.)

\begin{figure}[h]
\centering
%\vspace{-2mm}
\includegraphics[width=0.7\linewidth]{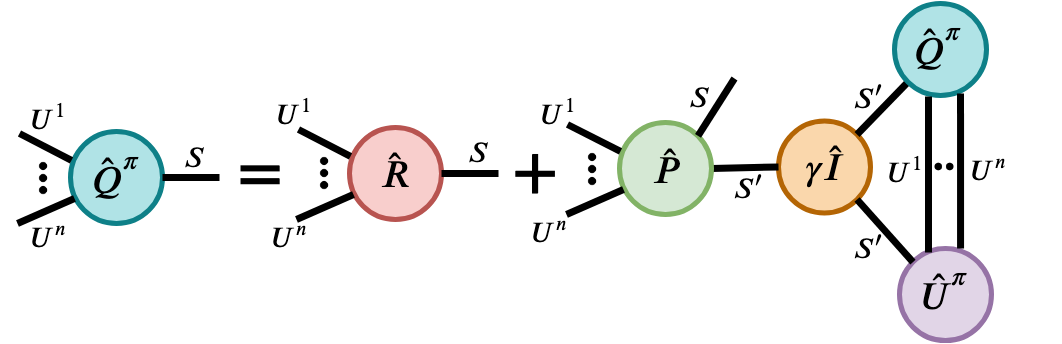}
\caption{\textbf{Tensorised Bellman Equation} for $n$ agents. There is an edge for each factor in product decomposition/agent $i \in \mathcal{A}$ in the corresponding nodes $\hat Q^\pi,\hat U^\pi, \hat R, \hat P$ with the index set $U^i$.\label{fig:tbell}}
%\vspace{-2mm}
\end{figure}

We are now ready to present the tensorised form of the Bellman equation shown in \cref{eq:bell}. \cref{fig:tbell} gives the equation where $\hat I$ is the identity tensor of size $|S|\times|S|\times|S|$. The dependence of the action-value tensor $\hat Q^\pi$ and the policy tensor $\hat U^\pi$ on the policy is denoted by superscripts $\pi$. The novel \textbf{Tensorised Bellman equation} provides a theoretically justified foundation for the approximation of the joint $Q$-function, and the subsequent analysis (Theorems 1-3) for learning using this approximation.

\subsection{\textsc{Tesseract} Algorithms}
\label{subsec:TessAlgos}
% \captionsetup[algorithm]{format=hang,singlelinecheck=false}
% \begin{wrapfigure}{L}{0.7\textwidth}
% \begin{minipage}{0.7\textwidth}
%   \begin{wrapfigure}{L}{0.5\textwidth}
%     \begin{minipage}{0.5\textwidth}
%       \begin{algorithm}[H]
%         \caption{assignment algorithm}
%         \begin{algorithmic}
%           \STATE i $\leftarrow$ j
%         \end{algorithmic}
%       \end{algorithm}
%     \end{minipage}
%   \end{wrapfigure}
\begin{wrapfigure}{R}{0.55\textwidth}
\vspace{-9mm}
\begin{minipage}{0.55\textwidth}
\begin{algorithm}[H]
	\caption{Model-based Tesseract\label{alg:model-based}}
	\begin{algorithmic}[1]
	    \STATE \mbox{Initialise rank $k$, $\pi = (\pi^i)_1^n$  and $\hat Q$}
		\STATE \mbox{Initialise model parameters  $\hat P,\hat R$}
		\STATE Learning rate $\leftarrow \alpha$,$\mathcal{D} \leftarrow \left\{ \right\}$ 
		\FOR{each episodic iteration i}
		\STATE Do episode rollout $\tau_i = \left\{(s_t,\mathbf{u}_t,r_t,s_{t+1})_{0}^L \right\}$ using $\pi$
		\STATE $\mathcal{D} \leftarrow \mathcal{D}\cup\left\{\tau_i \right\}$
		\STATE Update $\hat P,\hat R$ using CP-Decomposition on moments from $\mathcal{D}$
		\FOR{each internal iteration j}
% 		\STATE $\hat Q \gets \Pi_{k1+|S|k2}\mathcal{T}^\pi \hat Q$
		\STATE $\hat Q \gets \mathcal{T}^\pi \hat Q$
		\ENDFOR
		\STATE Improve $\pi$ using $\hat Q$% and your favourite algorithm.
		\ENDFOR
		\STATE Return $\pi, \hat Q$
	\end{algorithmic}
\end{algorithm}
\end{minipage}
\vspace{-3mm}
\end{wrapfigure}
% For any $k \in \mathbb{N}$ let $\mathcal{Q}_k \triangleq \{Q: Q\in \mathcal{Q} \land rank(Q(\cdot, s)) \leq k, \forall s \in S\}$. Given any policy $\pi$ we are interested in projecting $Q^\pi$ to $\mathcal{Q}_k$ using the projection operator $\Pi_k(\cdot) = \argmin_{Q \in \mathcal{Q}_k} ||\cdot-Q||_{\pi,F}$. where $||X||_{\pi,F} \triangleq \mathbb{E}_{s\sim\rho^\pi(s)}[||X(s)||_{F}]$ is the weighted Frobenius norm w.r.t.\ policy visitation over states. Thus a simple planning based algorithm for rank $k$ \textsc{Tesseract} would involve starting with an arbitrary $Q_0$ and successively applying the Bellman operator $\mathcal{T}^{\pi}$ and the projection operator $\Pi_k$ so that $Q_{t+1} = \Pi_k\mathcal{T}^{\pi}Q_t$. 
It turns out that, constraining the underlying tensors for dynamics and rewards ($\hat P, \hat R$) is sufficient to bound the CP-rank of $\hat Q$ (\cref{rankbQ}, \cite{pmlr-v139-mahajan21a}). From this insight, a model-based RL version for \textsc{Tesseract} can be constructed in \cref{alg:model-based} (reproduced here from \cite{pmlr-v139-mahajan21a}). The algorithm proceeds by estimating the underlying MDP dynamics using the sampled trajectories obtained by executing the behaviour policy $\pi = (\pi^i)_1^n$ satisfying \cref{thm:debound}. Specifically, we use a rank $k$ approximate CP-Decomposition to calculate the model dynamics $R, P$ as we show in \cref{analysis}. Next $\pi$ is evaluated using the estimated dynamics, which is followed by policy improvement, \cref{alg:model-based} gives the pseudocode for the model-based setting. The termination and policy improvement decisions in \cref{alg:model-based} admit a wide range of choices used in practice in the RL community. Example choices for internal iterations which broadly fall under approximate policy iteration include: 1) Fixing the number of applications of Bellman operator 2) Using norm of difference between consecutive Q estimates etc., similarly for policy improvement several options can be used like $\epsilon$-greedy (for Q derived policy), policy gradients (parametrized policy)~\cite{sutton2011reinforcement}
For large state spaces, where storage and planning using model parameters is computationally difficult, \cite{pmlr-v139-mahajan21a} provide a model free version of the approach, details of which can be found in \cref{app:mft} with a sample of empirical results from the original paper in \cref{sec:exps}. We next briefly revisit the main theoretical results from \textsc{Tesseract}. Additional related works can be found in \cref{sec:relw}.

\section{Analysis}

\label{analysis}
Under mild assumptions on the underlying dynamics of the FAS problem (see \cref{app:assump} for these along with additional constants appearing in \cref{thm:debound}), we have that the following results hold, where $k_1, k_2$ are upper bounds on CP rank of the reward and transition tensor respectively (\textbf{proofs for which can be found in the original text, \cite{pmlr-v139-mahajan21a}}):

\begin{theorem}
\label{rankbQ}
For a finite FAS the action-value tensor satisfies $rank(\hat Q^\pi(s))\leq k_1+k_2|S|,\forall s \in S, \forall \pi$.
\end{theorem}

% \cref{rankbQ} implies that for approximations with enough factors, policy evaluation converges:
% \begin{corollary}
% \label{cor:suf_rank}
% For all $k\geq k_1+k_2|S|$, the procedure $Q_{t+1}\leftarrow \Pi_{k}\mathcal{T}^{\pi}Q_{t}$ converges to $Q^\pi$ for all $Q_0,\pi$.
% \end{corollary}

% \cref{cor:suf_rank} is especially useful for the case of M-POMDP and M-ROMDP with $|Z| >> |S|$, i.e., where the intrinsic state space dimensionality is small in comparison to the dimensionality of the observations. In these cases the Tensorised Bellman equation \cref{fig:tbell} can be augmented by padding the transition tensor $\hat P$ with the observation matrix and the lower bound in \cref{cor:suf_rank} can be improved using the intrinsic state dimensionality.

The next PAC result empowers \cref{alg:model-based} by lower bounding the number of samples required to infer the reward and state transition dynamics for finite MDPs with high probability using sufficient approximate rank $k \geq k_1, k_2$: 

\begin{theorem}[Model based estimation of $\hat R, \hat P$ error bounds]
\label{thm:debound}
Given any $\epsilon>0, 1>\delta>0$, for a policy $\pi$ with the policy tensor satisfying $\pi(\mathbf{u}|s)\geq \Delta$, where
\begin{align}
% \label{eq:polcon}
\nonumber
\Delta = \max_s 
\frac{C_1 \mu_{s}^6 k^5 (w_{s}^{\text{max}})^4 \log(|U|)^4 \log(3k||R(s)||_{F}/\epsilon)}
{|U|^{n/2} (w_{s}^{\text{min}})^4}  
\end{align}
and $C_1$ is a problem dependent positive constant. There exists $N_0$ which is $O(|U|^{\frac{n}{2}})$ and polynomial in $\frac{1}{\delta},\frac{1}{\epsilon}, k$ and relevant spectral properties of the underlying MDP dynamics such that for samples $\geq N_0$, we can compute the estimates $\bar R(s), \bar P(s,s')$ such that w.p. $\geq 1-\delta$, $||\bar{R}(s)-\hat R(s)||_F\leq \epsilon, ||\bar{P}(s,s')-\hat P(s,s')||_F\leq \epsilon, \forall s,s' \in S$.
\end{theorem}

\cref{thm:debound} gives the relation between the order of the number of samples required to estimate dynamics and the tolerance for approximation. \cref{thm:debound} states that aside from allowing efficient PAC learning of the reward and transition dynamics of the factored action-space MDP, \cref{alg:model-based} requires only $O(|U|^{\frac{n}{2}})$ to do so, which is a vanishing fraction of $|U|^n$, the total number of joint actions in any given state. This also hints at why a tensor based approximation of the $Q$-function helps with sample efficiency. Methods that do not use the tensor structure typically use $O(|U|^n)$ samples. The bound is also useful for off-policy scenarios, where only the behaviour policy needs to satisfy the bound.
Given the result in \cref{thm:debound}, it is natural to ask what is the error associated with computing the action-values of a policy using the estimated transition and reward dynamics. This is addressed in the next result, 
% but first we present a lemma bounding the total variation distance between the estimated and true transition distributions: 
% \begin{lemma}
% \label{tvbound}
% For transition tensor estimates satisfying $||\bar{P}(s,s')-\hat P(s,s')||_F\leq \epsilon$, we have for any given state-action pair $(s,a)$, the distribution over the next states follows: $TV(P'(\cdot|s,a),P(\cdot|s,a))\leq \frac{1}{2}(|1-f|+f|S|\epsilon)$ where $\frac{1}{1+\epsilon|S|}\leq f \leq\frac{1}{1-\epsilon|S|}$, where $TV$ is the \textit{total variation} distance. Similarly for any policy $\pi$, $TV(\bar P_{\pi}(\cdot|s),P_{\pi}(\cdot|s)), TV(\bar P_{\pi}(s',a'|s),P_{\pi}(s',a'|s))\leq \frac{1}{2}(|1-f|+f|S|\epsilon)$ 
% \end{lemma}
which bounds the error of model-based evaluation using approximate dynamics in \cref{thm:q_err}. The first component on the RHS of the upper bound comes from the tensor analysis of the transition dynamics, whereas the second component can be attributed to error propagation for the rewards.
% \vspace{5mm}
\begin{theorem}[Error bound on policy evaluation]
\label{thm:q_err}
Given a behaviour policy $\pi_b$ satisfying the conditions in \cref{thm:debound} and executed for steps $\geq N_0$, for any policy $\pi$ the model based policy evaluation $Q_{\bar P,\bar R}^\pi$ satisfies:
% \vspace{-5pt}
\begin{align}
|Q_{P,R}^\pi(s,a) - Q_{\bar P,\bar R}^\pi(s,a)|\leq &(|1-f|+f|S|\epsilon)\frac{\gamma}{2(1-\gamma)^2}+ \frac{\epsilon}{1-\gamma}, \forall (s,a)\in S\times U^n
\end{align} where $f$ is $\frac{1}{1+\epsilon|S|}\leq f \leq\frac{1}{1-\epsilon|S|}$
\end{theorem}
% Additional theoretical discussion can be found in \cref{app:atd}
%%%%%%%%%%%%%%%%%%%%%%%%%%%%%%%%%%%%%%%%%%%%%%%%%%%%%%%

% \input{sections/experiments.tex}	
% \input{sections/related_work.tex}	
\clearpage
\newpage
\section{Conclusions \& Future Work}
In this extended abstract we discussed the main ideas introduced in \textsc{Tesseract}~\citep{pmlr-v139-mahajan21a}, a novel framework utilising the insight that the action value function for RL problems with factored action space can be seen as a tensor. \textsc{Tesseract} provides a means for developing new sample efficient algorithms and obtain essential guarantees about convergence and recovery of the underlying dynamics. We also revisited the main theoretical results of \textsc{Tesseract}. 
There are several interesting open questions to address in future work, such as convergence and error analysis for rank insufficient approximation, and analysis of the learning framework under different types of tensor decompositions like Tucker and tensor-train \citep{kolda2009tensor}.

\bibliographystyle{unsrtnat}
\bibliography{tesseract}

\begin{thebibliography}{44}
\providecommand{\natexlab}[1]{#1}
\providecommand{\url}[1]{\texttt{#1}}
\expandafter\ifx\csname urlstyle\endcsname\relax
  \providecommand{\doi}[1]{doi: #1}\else
  \providecommand{\doi}{doi: \begingroup \urlstyle{rm}\Url}\fi

\bibitem[Mahajan et~al.(2021)Mahajan, Samvelyan, Mao, Makoviychuk, Garg,
  Kossaifi, Whiteson, Zhu, and Anandkumar]{pmlr-v139-mahajan21a}
Anuj Mahajan, Mikayel Samvelyan, Lei Mao, Viktor Makoviychuk, Animesh Garg,
  Jean Kossaifi, Shimon Whiteson, Yuke Zhu, and Animashree Anandkumar.
\newblock Tesseract: Tensorised actors for multi-agent reinforcement learning.
\newblock In \emph{Proceedings of the 38th International Conference on Machine
  Learning}, volume 139, pages 7301--7312. PMLR, 2021.
\newblock URL \url{https://proceedings.mlr.press/v139/mahajan21a.html}.

\bibitem[Silver et~al.(2017)Silver, Schrittwieser, Simonyan, Antonoglou, Huang,
  Guez, Hubert, Baker, Lai, Bolton, et~al.]{silver2017mastering}
David Silver, Julian Schrittwieser, Karen Simonyan, Ioannis Antonoglou, Aja
  Huang, Arthur Guez, Thomas Hubert, Lucas Baker, Matthew Lai, Adrian Bolton,
  et~al.
\newblock Mastering the game of go without human knowledge.
\newblock \emph{nature}, 550\penalty0 (7676):\penalty0 354--359, 2017.

\bibitem[Vinyals et~al.(2019)Vinyals, Babuschkin, Czarnecki, Mathieu, Dudzik,
  Chung, Choi, Powell, Ewalds, Georgiev, et~al.]{vinyals2019grandmaster}
Oriol Vinyals, Igor Babuschkin, Wojciech~M Czarnecki, Micha{\"e}l Mathieu,
  Andrew Dudzik, Junyoung Chung, David~H Choi, Richard Powell, Timo Ewalds,
  Petko Georgiev, et~al.
\newblock Grandmaster level in starcraft ii using multi-agent reinforcement
  learning.
\newblock \emph{Nature}, 575\penalty0 (7782):\penalty0 350--354, 2019.

\bibitem[DeepMind-OEL et~al.(2021)DeepMind-OEL, Stooke, Mahajan, Barros, Deck,
  Bauer, Sygnowski, Trebacz, Jaderberg, Mathieu, McAleese, Bradley-Schmieg,
  Wong, Porcel, Raileanu, Hughes-Fitt, Dalibard, and Czarnecki]{oel2021}
DeepMind-OEL, Adam Stooke, Anuj Mahajan, Catarina Barros, Charlie Deck, Jakob
  Bauer, Jakub Sygnowski, Maja Trebacz, Max Jaderberg, Michael Mathieu, Nat
  McAleese, Nathalie Bradley-Schmieg, Nathaniel Wong, Nicolas Porcel, Roberta
  Raileanu, Steph Hughes-Fitt, Valentin Dalibard, and Wojciech~Marian
  Czarnecki.
\newblock Open-ended learning leads to generally capable agents.
\newblock \emph{arXiv preprint arXiv:2107.12808}, 2021.

\bibitem[Bu{\c{s}}oniu et~al.(2010)Bu{\c{s}}oniu, Babu{\v{s}}ka, and
  De~Schutter]{bucsoniu2010multi}
Lucian Bu{\c{s}}oniu, Robert Babu{\v{s}}ka, and Bart De~Schutter.
\newblock Multi-agent reinforcement learning: An overview.
\newblock In \emph{Innovations in multi-agent systems and applications-1},
  pages 183--221. Springer, 2010.

\bibitem[Sutton and Barto(2011)]{sutton2011reinforcement}
Richard~S Sutton and Andrew~G Barto.
\newblock Reinforcement learning: An introduction.
\newblock 2011.

\bibitem[Kolda and Bader(2009)]{kolda2009tensor}
Tamara~G Kolda and Brett~W Bader.
\newblock Tensor decompositions and applications.
\newblock \emph{SIAM review}, 51\penalty0 (3):\penalty0 455--500, 2009.

\bibitem[Janzamin et~al.(2020)Janzamin, Ge, Kossaifi, and
  Anandkumar]{janzamin2020spectral}
Majid Janzamin, Rong Ge, Jean Kossaifi, and Anima Anandkumar.
\newblock Spectral learning on matrices and tensors.
\newblock \emph{arXiv preprint arXiv:2004.07984}, 2020.

\bibitem[Barendregt(1984)]{barendregt1984introduction}
Henk~P Barendregt.
\newblock Introduction to lambda calculus.
\newblock 1984.

\bibitem[Greensmith et~al.(2004)Greensmith, Bartlett, and
  Baxter]{greensmith2004variance}
Evan Greensmith, Peter~L Bartlett, and Jonathan Baxter.
\newblock Variance reduction techniques for gradient estimates in reinforcement
  learning.
\newblock \emph{Journal of Machine Learning Research}, 5\penalty0 (9), 2004.

\bibitem[Zhao et~al.(2016)Zhao, Niu, Xie, Yang, and
  Sugiyama]{zhao2016regularized}
Tingting Zhao, Gang Niu, Ning Xie, Jucheng Yang, and Masashi Sugiyama.
\newblock Regularized policy gradients: direct variance reduction in policy
  gradient estimation.
\newblock In \emph{Asian Conference on Machine Learning}, pages 333--348. PMLR,
  2016.

\bibitem[Sutton(1988)]{sutton1988learning}
Richard~S Sutton.
\newblock Learning to predict by the methods of temporal differences.
\newblock \emph{Machine learning}, 3\penalty0 (1):\penalty0 9--44, 1988.

\bibitem[Son et~al.(2019)Son, Kim, Kang, Hostallero, and Yi]{son2019qtran}
Kyunghwan Son, Daewoo Kim, Wan~Ju Kang, David~Earl Hostallero, and Yung Yi.
\newblock Qtran: Learning to factorize with transformation for cooperative
  multi-agent reinforcement learning.
\newblock \emph{arXiv preprint arXiv:1905.05408}, 2019.

\bibitem[Samvelyan et~al.(2019)Samvelyan, Rashid, de~Witt, Farquhar, Nardelli,
  Rudner, Hung, Torr, Foerster, and Whiteson]{samvelyan2019starcraft}
Mikayel Samvelyan, Tabish Rashid, Christian~Schroeder de~Witt, Gregory
  Farquhar, Nantas Nardelli, Tim~GJ Rudner, Chia-Man Hung, Philip~HS Torr,
  Jakob Foerster, and Shimon Whiteson.
\newblock {The} {StarCraft} {Multi-Agent} {Challenge}.
\newblock In \emph{Proceedings of the 18th International Conference on
  Autonomous Agents and MultiAgent Systems}, 2019.

\bibitem[Rashid et~al.(2018)Rashid, Samvelyan, de~Witt, Farquhar, Foerster, and
  Whiteson]{rashid2018qmix}
Tabish Rashid, Mikayel Samvelyan, Christian~Schroeder de~Witt, Gregory
  Farquhar, Jakob Foerster, and Shimon Whiteson.
\newblock {QMIX}: {Monotonic} {Value} {Function} {Factorisation} for {Deep}
  {Multi-Agent} {Reinforcement} {Learning}.
\newblock In \emph{Proceedings of the 35th International Conference on Machine
  Learning}, pages 4295--4304, 2018.

\bibitem[Sunehag et~al.(2017)Sunehag, Lever, Gruslys, Czarnecki, Zambaldi,
  Jaderberg, Lanctot, Sonnerat, Leibo, Tuyls, and
  Graepel]{sunehag_value-decomposition_2017}
Peter Sunehag, Guy Lever, Audrunas Gruslys, Wojciech~Marian Czarnecki, Vinicius
  Zambaldi, Max Jaderberg, Marc Lanctot, Nicolas Sonnerat, Joel~Z. Leibo, Karl
  Tuyls, and Thore Graepel.
\newblock Value-{Decomposition} {Networks} {For} {Cooperative} {Multi}-{Agent}
  {Learning} {Based} {On} {Team} {Reward}.
\newblock In \emph{Proceedings of the 17th International Conference on
  Autonomous Agents and Multiagent Systems}, 2017.

\bibitem[Chen et~al.(2018)Chen, Zhou, Wen, Yang, Su, Zhang, Zhang, Wang, and
  Liu]{chen2018factorized}
Yong Chen, Ming Zhou, Ying Wen, Yaodong Yang, Yufeng Su, Weinan Zhang, Dell
  Zhang, Jun Wang, and Han Liu.
\newblock Factorized q-learning for large-scale multi-agent systems.
\newblock \emph{arXiv preprint arXiv:1809.03738}, 2018.

\bibitem[Tan(1993)]{tan_multi-agent_1993}
Ming Tan.
\newblock Multi-agent reinforcement learning: {Independent} vs. cooperative
  agents.
\newblock In \emph{Proceedings of the Tenth International Conference on Machine
  Learning}, pages 330--337, 1993.

\bibitem[Mahajan et~al.(2019)Mahajan, Rashid, Samvelyan, and
  Whiteson]{mahajan2019maven}
Anuj Mahajan, Tabish Rashid, Mikayel Samvelyan, and Shimon Whiteson.
\newblock Maven: Multi-agent variational exploration.
\newblock In \emph{Advances in Neural Information Processing Systems}, pages
  7611--7622, 2019.

\bibitem[Guestrin et~al.(2002{\natexlab{a}})Guestrin, Lagoudakis, and
  Parr]{guestrin2002coordinated}
Carlos Guestrin, Michail Lagoudakis, and Ronald Parr.
\newblock Coordinated reinforcement learning.
\newblock In \emph{ICML}, volume~2, pages 227--234. Citeseer,
  2002{\natexlab{a}}.

\bibitem[Guestrin et~al.(2002{\natexlab{b}})Guestrin, Venkataraman, and
  Koller]{guestrin2002context}
Carlos Guestrin, Shobha Venkataraman, and Daphne Koller.
\newblock Context-specific multiagent coordination and planning with factored
  mdps.
\newblock In \emph{AAAI/IAAI}, pages 253--259, 2002{\natexlab{b}}.

\bibitem[Bargiacchi et~al.(2018)Bargiacchi, Verstraeten, Roijers, Now{\'e}, and
  Hasselt]{bargiacchi2018learning}
Eugenio Bargiacchi, Timothy Verstraeten, Diederik Roijers, Ann Now{\'e}, and
  Hado Hasselt.
\newblock Learning to coordinate with coordination graphs in repeated
  single-stage multi-agent decision problems.
\newblock In \emph{International conference on machine learning}, pages
  482--490, 2018.

\bibitem[Gupta et~al.(2020)Gupta, Mahajan, Peng, B{\"o}hmer, and
  Whiteson]{gupta2020uneven}
Tarun Gupta, Anuj Mahajan, Bei Peng, Wendelin B{\"o}hmer, and Shimon Whiteson.
\newblock Uneven: Universal value exploration for multi-agent reinforcement
  learning.
\newblock \emph{arXiv preprint arXiv:2010.02974}, 2020.

\bibitem[Yang et~al.(2020)Yang, Hao, Liao, Shao, Chen, Liu, and
  Tang]{Yang2020QattenAG}
Yaodong Yang, Jianye Hao, Ben~Lu Liao, Kun Shao, Guangyong Chen, Wulong Liu,
  and Hongyao Tang.
\newblock Qatten: A general framework for cooperative multiagent reinforcement
  learning.
\newblock \emph{ArXiv}, abs/2002.03939, 2020.

\bibitem[Wang et~al.(2020)Wang, Gupta, Mahajan, Peng, Whiteson, and
  Zhang]{wang2020rode}
Tonghan Wang, Tarun Gupta, Anuj Mahajan, Bei Peng, Shimon Whiteson, and
  Chongjie Zhang.
\newblock Rode: Learning roles to decompose multi-agent tasks.
\newblock \emph{arXiv preprint arXiv:2010.01523}, 2020.

\bibitem[Lowe et~al.(2017)Lowe, Wu, Tamar, Harb, Abbeel, and
  Mordatch]{lowe2017multi}
Ryan Lowe, Yi~Wu, Aviv Tamar, Jean Harb, OpenAI~Pieter Abbeel, and Igor
  Mordatch.
\newblock Multi-agent actor-critic for mixed cooperative-competitive
  environments.
\newblock In \emph{Advances in Neural Information Processing Systems}, pages
  6379--6390, 2017.

\bibitem[Foerster et~al.(2018)Foerster, Farquhar, Afouras, Nardelli, and
  Whiteson]{foerster2018counterfactual}
Jakob~N Foerster, Gregory Farquhar, Triantafyllos Afouras, Nantas Nardelli, and
  Shimon Whiteson.
\newblock Counterfactual multi-agent policy gradients.
\newblock In \emph{Thirty-Second AAAI Conference on Artificial Intelligence},
  2018.

\bibitem[Mahajan and Tulabandhula(2017{\natexlab{a}})]{mahajan2017symmetryde}
Anuj Mahajan and Theja Tulabandhula.
\newblock Symmetry detection and exploitation for function approximation in
  deep rl.
\newblock In \emph{Proceedings of the 16th Conference on Autonomous Agents and
  MultiAgent Systems}, pages 1619--1621, 2017{\natexlab{a}}.

\bibitem[Mahajan and Tulabandhula(2017{\natexlab{b}})]{mahajan2017symmetryl}
Anuj Mahajan and Theja Tulabandhula.
\newblock Symmetry learning for function approximation in reinforcement
  learning.
\newblock \emph{arXiv preprint arXiv:1706.02999}, 2017{\natexlab{b}}.

\bibitem[Kakade(2003)]{kakade2003sample}
Sham~Machandranath Kakade.
\newblock \emph{On the sample complexity of reinforcement learning}.
\newblock PhD thesis, UCL (University College London), 2003.

\bibitem[Lattimore et~al.(2013)Lattimore, Hutter, and
  Sunehag]{lattimore2013sample}
Tor Lattimore, Marcus Hutter, and Peter Sunehag.
\newblock The sample-complexity of general reinforcement learning.
\newblock In \emph{International Conference on Machine Learning}, pages 28--36.
  PMLR, 2013.

\bibitem[Anandkumar et~al.(2014)Anandkumar, Ge, Hsu, Kakade, and
  Telgarsky]{anandkumar2014tensor}
Animashree Anandkumar, Rong Ge, Daniel Hsu, Sham~M Kakade, and Matus Telgarsky.
\newblock Tensor decompositions for learning latent variable models.
\newblock \emph{Journal of Machine Learning Research}, 15:\penalty0 2773--2832,
  2014.

\bibitem[{Sidiropoulos} et~al.(2017){Sidiropoulos}, {De Lathauwer}, {Fu},
  {Huang}, {Papalexakis}, and {Faloutsos}]{sidiropoulos2017tensor}
N.~D. {Sidiropoulos}, L.~{De Lathauwer}, X.~{Fu}, K.~{Huang}, E.~E.
  {Papalexakis}, and C.~{Faloutsos}.
\newblock Tensor decomposition for signal processing and machine learning.
\newblock \emph{IEEE Transactions on Signal Processing}, 65\penalty0
  (13):\penalty0 3551--3582, 2017.

\bibitem[Panagakis et~al.(2021)Panagakis, Kossaifi, Chrysos, Oldfield,
  Nicolaou, Anandkumar, and Zafeiriou]{panagakis2021tensor}
Yannis Panagakis, Jean Kossaifi, Grigorios~G Chrysos, James Oldfield, Mihalis~A
  Nicolaou, Anima Anandkumar, and Stefanos Zafeiriou.
\newblock Tensor methods in computer vision and deep learning.
\newblock \emph{Proceedings of the IEEE}, 109\penalty0 (5):\penalty0 863--890,
  2021.

\bibitem[Cohen et~al.(2016)Cohen, Sharir, and Shashua]{cohen2016expressive}
Nadav Cohen, Or~Sharir, and Amnon Shashua.
\newblock On the expressive power of deep learning: A tensor analysis.
\newblock In \emph{Conference on Learning Theory}, pages 698--728, 2016.

\bibitem[Cheng et~al.(2017)Cheng, Wang, Zhou, and Zhang]{cheng2017survey}
Yu~Cheng, Duo Wang, Pan Zhou, and Tao Zhang.
\newblock A survey of model compression and acceleration for deep neural
  networks.
\newblock \emph{arXiv preprint arXiv:1710.09282}, 2017.

\bibitem[Cichocki et~al.(2017)Cichocki, Phan, Zhao, Lee, Oseledets, Sugiyama,
  Mandic, et~al.]{cichocki2017tensor}
Andrzej Cichocki, Anh-Huy Phan, Qibin Zhao, Namgil Lee, Ivan Oseledets, Masashi
  Sugiyama, Danilo~P Mandic, et~al.
\newblock Tensor networks for dimensionality reduction and large-scale
  optimization: Part 2 applications and future perspectives.
\newblock \emph{Foundations and Trends{\textregistered} in Machine Learning},
  9\penalty0 (6):\penalty0 431--673, 2017.

\bibitem[Kossaifi et~al.(2019)Kossaifi, Bulat, Tzimiropoulos, and
  Pantic]{t_net}
Jean Kossaifi, Adrian Bulat, Georgios Tzimiropoulos, and Maja Pantic.
\newblock T-net: Parametrizing fully convolutional nets with a single
  high-order tensor.
\newblock In \emph{The IEEE Conference on Computer Vision and Pattern
  Recognition (CVPR)}, June 2019.

\bibitem[Kossaifi et~al.(2020)Kossaifi, Toisoul, Bulat, Panagakis, Hospedales,
  and Pantic]{kossaifi2019efficient}
Jean Kossaifi, Antoine Toisoul, Adrian Bulat, Yannis Panagakis, Timothy
  Hospedales, and Maja Pantic.
\newblock Factorized higher-order cnns with an application to spatio-temporal
  emotion estimation.
\newblock In \emph{IEEE CVPR}, 2020.

\bibitem[Bulat et~al.(2020)Bulat, Kossaifi, Tzimiropoulos, and
  Pantic]{bulat2019incremental}
Adrian Bulat, Jean Kossaifi, Georgios Tzimiropoulos, and Maja Pantic.
\newblock Incremental multi-domain learning with network latent tensor
  factorization.
\newblock 2020.

\bibitem[Azizzadenesheli et~al.(2016)Azizzadenesheli, Lazaric, and
  Anandkumar]{azizzadenesheli2016reinforcement}
Kamyar Azizzadenesheli, Alessandro Lazaric, and Animashree Anandkumar.
\newblock Reinforcement learning in rich-observation mdps using spectral
  methods.
\newblock \emph{arXiv preprint arXiv:1611.03907}, 2016.

\bibitem[Azizzadenesheli(2019)]{azizzadenesheli2019reinforcement}
Kamyar Azizzadenesheli.
\newblock \emph{Reinforcement Learning in Structured and Partially Observable
  Environments}.
\newblock PhD thesis, UC Irvine, 2019.

\bibitem[Krishnamurthy et~al.(2016)Krishnamurthy, Agarwal, and
  Langford]{krishnamurthy2016pac}
Akshay Krishnamurthy, Alekh Agarwal, and John Langford.
\newblock Pac reinforcement learning with rich observations.
\newblock In \emph{Advances in Neural Information Processing Systems}, pages
  1840--1848, 2016.

\bibitem[Bromuri(2012)]{bromuri2012tensor}
Stefano Bromuri.
\newblock A tensor factorization approach to generalization in multi-agent
  reinforcement learning.
\newblock In \emph{2012 IEEE/WIC/ACM International Conferences on Web
  Intelligence and Intelligent Agent Technology}, volume~2, pages 274--281.
  IEEE, 2012.

\end{thebibliography}
\clearpage
\newpage

\appendix

\section{Model free Tesseract}
\label{app:mft}
\begin{figure}
    \centering
    %\vspace{-0.3cm}
    \includegraphics[width=0.7\linewidth]{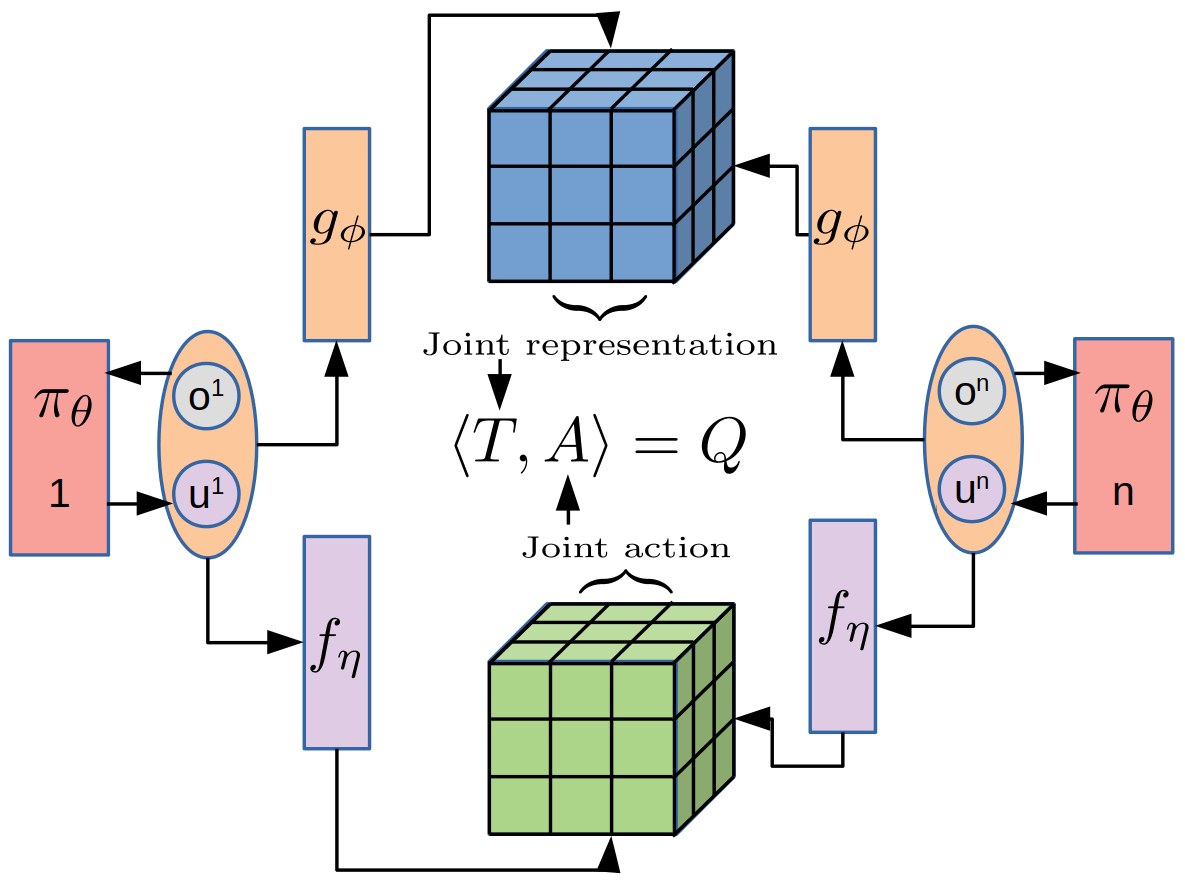}
    \caption{Tesseract architecture reproduced from original paper \cite{pmlr-v139-mahajan21a} \label{fig:tnet}}
    %\vspace{-0.5cm}
\end{figure}

For large state spaces where storage and planning using model parameters is computationally difficult (they are $\mathcal{O}(kn|U||S|^2)$ in number), 
% \sw{Be explicit about what the space savings are ie what is the dimensionality of the objects that need to be stored.  I'm not sure about the computational argument as once you've learned a model you can do as much or as little planning with it as your budget allows.} 
~\cite{pmlr-v139-mahajan21a} propose a model-free approach using a deep network where the rank constraint on the $Q$-function is directly embedded into the network architecture. \cref{fig:tnet} gives the general network architecture for this approach and \cref{alg:model-free} the associated pseudo-code. Each agent in \cref{fig:tnet} has a policy network parameterized by $\theta$ which is used to take actions in a decentralised manner. 
% \sw{So at the same time that you switched from model-based to model-free, you switch from an MDP to a decentralised partially observable setting, but without mentioning it or explaining the connection between the change in setting and the change in algorithm.  This is quite confusing.} 
The observations of the individual agents along with the actions are fed through representation function $g_\phi$ whose output is a set of $k$ unit vectors of dimensionality $|U|$ corresponding to each rank. 
% \sw{It sounds like this is a Dec-MDP not a Dec-POMDP, otherwise the join observation might not disambiguate the state.} 
The output $g_{\phi,r}(s^i)$ corresponding to each agent $i$ for factor $r$ can be seen as an action-wise contribution to the joint utility from the agent corresponding to that factor. The joint utility here is a product over individual agent utilities. For partially observable settings, an additional RNN layer can be used to summarise agent trajectories. The joint action-value estimate of the tensor $\hat Q(s)$ by the centralized critic is: 
%\vspace{-0.3cm}
\begin{align}
\label{eq:cpa}
\hat Q(s) \approx T = \sum_{r=1}^k w_r\otimes^n g_{\phi,r}(s^i) ,i 
\in \{1..n\}, \vspace{-5pt}
\end{align}
% \sw{You haven't mentioned CTDE since the intro or explained that learning a centralised critic that is discarded upon deployment is a standard way to exploit CTDE.}
where the weights $w_r$ are learnable parameters exclusive to the centralized learner. In the case of value based methods where the policy is implicitly derived from utilities, the policy parameters $\theta$ are merged with $\phi$. The network architecture is agnostic to the type of the action space (discrete/continuous) and the action-value corresponding to a particular joint-action $(u^1..u^n)$ is the inner product $\langle T, A \rangle$ where $A = \otimes^n u^i$ (This reduces to indexing using joint action in \cref{eq:cpa} for discrete spaces). More representational capacity can be added to the network by creating an abstract representation for actions using $f_\eta$, which can be any arbitrary monotonic function (parametrised by $\eta$) of vector output of size $m \geq |U|$ and preserves relative order of utilities across actions; this ensures that the optimal policy is learnt as long as it belongs to the hypothesis space. 
% \sw{What does this condition ensure?  Support the claim.} 
In this case $A = \otimes^n f_\eta(u^i)$ and the agents also carry a copy of $f_\eta$ during the execution phase. Furthermore, the inner product $\langle T, A \rangle$ can be computed efficiently using the property $$\langle T, A \rangle = \sum_{r=1}^k w_r\prod_1^n \langle f_\eta(u^i)g_{\phi,r}(s^i) \rangle ,i \in \{1..n\}$$ which is $O(nkm)$ whereas a naive approach involving computation of the tensors first would be $O(km^n)$. Training the Tesseract-based $Q$-network involves minimising the squared TD loss \cite{sutton2011reinforcement}:
\begin{align}
\mathcal{L}_{TD}(\phi,\eta) = \mathbb{E}_{\pi}[(&Q(\mathbf{u}_t,s_t; \phi,\eta)- [r(\mathbf{u}_t,s_t)+\gamma Q(\mathbf{u}_{t+1},s_{t+1}; \phi^-,\eta^-)])^2],
\end{align}
where $\phi^-,\eta^-$ are target parameters. Policy updates involve gradient ascent w.r.t.\ to the policy parameters $\theta$ on the objective $\mathcal{J}_\theta=\int_{S} \rho^\pi(s)\int_{\mathbf{U}}\pi_\theta(\mathbf{u|s})Q^{\pi}(s,\mathbf{u}) d\mathbf{u}ds$. More sophisticated targets can be used to reduce the policy gradient variance \citep{greensmith2004variance, zhao2016regularized} and propagate rewards efficiently \citep{sutton1988learning}. Note that 
\cref{alg:model-free} does not require the individual-global maximisation principle \citep{son2019qtran} typically assumed by value-based MARL methods in the CTDE setting, as it is an actor-critic method. 
% Tesseract also supports learning different factors \cref{eq:cpa} in separate episodic phases to ensure \textit{committed exploration}. \sw{I don't understand this sentence.} 
In general, any form of function approximation and compatible model-free approach can be interleaved with Tesseract by appropriate use of the projection function $\Pi_k$. 

% \vspace{-.2cm}
\begin{algorithm}[h!]
    \caption{Model-free Tesseract\label{alg:model-free}}
    \begin{algorithmic}[1]
	    \STATE \mbox{Initialise rank $k$, parameter vectors $\theta, \phi, \eta$}
		\STATE Learning rate $\leftarrow \alpha$,$\mathcal{D} \leftarrow \left\{ \right\}$ 
		\FOR{each episodic iteration i}
		\STATE Do episode rollout $\tau_i = \left\{(s_t,\mathbf{u}_t,r_t,s_{t+1})_{0}^L \right\}$ using $\pi_\theta$
		\STATE $\mathcal{D} \leftarrow \mathcal{D}\cup\left\{\tau_i \right\}$
		\STATE Sample batch $\mathcal{B} \subseteq \mathcal{D}$.
		\STATE Compute empirical estimates for $\mathcal{L}_{TD}, \mathcal{J}_\theta$
		\STATE $\phi \leftarrow \phi - \alpha \nabla_\phi \mathcal{L}_{TD}$ (Rank $k$ projection step)
		\STATE $\eta \leftarrow \eta - \alpha \nabla_\eta \mathcal{L}_{TD}$ (Action representation update)
		\STATE $\theta \leftarrow \theta + \alpha \nabla_\theta \mathcal{J}_\theta$ (Policy update)
		\ENDFOR
		\STATE Return $\pi, \hat Q$
	\end{algorithmic}
\end{algorithm}

\section{Assumptions used for analysis}
\label{app:assump}
The following assumptions are used in \textsc{Tesseract} \citep{pmlr-v139-mahajan21a} which we reproduce here for reference:
\begin{assumption}
\label{a1}
For the given MMDP $G=\left\langle S,U,P,r,n,\gamma\right\rangle$, the reward tensor $\hat R(s),\forall s\in S$ has bounded rank $k_1\in \mathbb{N}$. 
\end{assumption}

Intuitively, a small $k_1$ in \cref{a1} implies that the reward is dependent only on a small number of intrinsic factors characterising the actions.

\begin{assumption}
\label{a2}
For the given MMDP $G=\left\langle S,U,P,r,n,\gamma\right\rangle$, the transition tensor $\hat P(s,s'),\forall s,s'\in S$ has bounded rank $k_2\in \mathbb{N}$.
\end{assumption}

Intuitively a small $k_2$ in \cref{a2} implies that only a small number of intrinsic factors characterising the actions lead to meaningful change in the joint state.
\cref{a1}{-2} always hold for a finite MMDP as CP-rank is upper bounded by $\Pi_{j=1}^n |U_j|$, where $U_j$ are the action sets.
\begin{assumption}
\label{a3}
The underlying MMDP is ergodic for any policy $\pi$ so that there is a stationary distribution $\rho^\pi$.
\end{assumption}

Next, we define coherence parameters, which are quantities of interest for our theoretical results: for reward decomposition $\hat R(s) = \sum_r w_{r,s}\otimes^n v_{r,i,s}$, let $\mu_{s} = \sqrt{n}\max_{i,r,j}|v_{r,i,s}(j)|$, $w_{s}^{\text{max}} = \max_{i,r}w_{r,s}$, $w_{s}^{\text{min}} = \min_{i,r}w_{r,s}$. Similarly define the corresponding quantities for $\mu_{s,s'}, w_{s,s'}^{\text{max}},w_{s,s'}^{\text{min}}$ for transition tensors $\hat P(s,s')$. A low coherence implies that the tensor's mass is evenly spread and helps bound the possibility of never seeing an entry with very high mass (large absolute value of an entry).
\section{Experiments}
\label{sec:exps}

In this section we reproduce the main empirical results for \textsc{Tesseract} on the StarCraft domain for the decentralised multi agent scenario. Complete experimental analysis and setup details can be found in the original paper. The experiments use the model-free version of \textsc{Tesseract} (\cref{alg:model-free}).

\begin{figure*}[h]
	\centering
	\subfigure[3s5z \textbf{Easy}]{
		\includegraphics[width=0.325\linewidth]{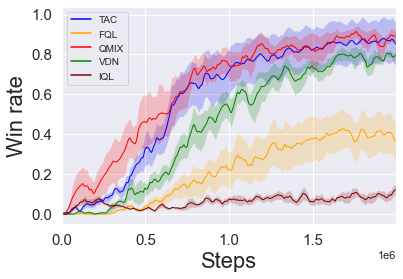}\label{fig:3s5z_smac}}
	\subfigure[2s\_vs\_1sc \textbf{Easy}]{
		\includegraphics[width=0.325\linewidth]{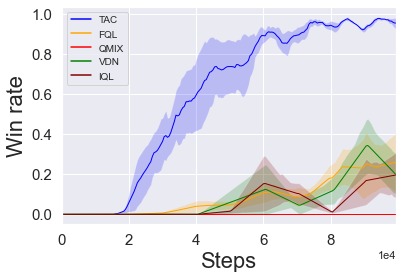}\label{fig:2s_vs_1sc}}
	\subfigure[2c\_vs\_64zg \textbf{Hard}]{
		\includegraphics[width=0.325\linewidth]{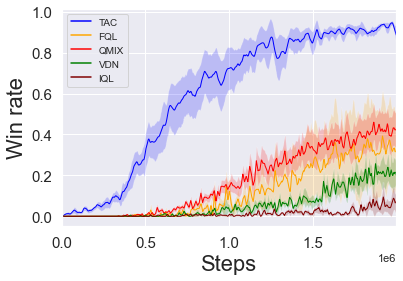}\label{fig:2c_vs_64z}}
% 	\vspace{-2mm}
	\subfigure[5m\_vs\_6m \textbf{Hard}]{
		\includegraphics[width=0.325\linewidth]{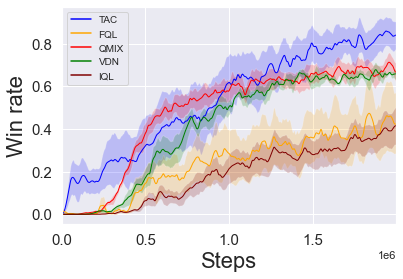}\label{fig:5m_vs_6m}}
	\subfigure[MMM2 \textbf{Super Hard}]{
		\includegraphics[width=0.325\linewidth]{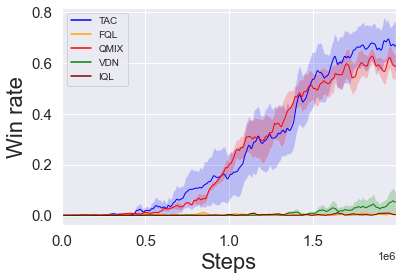}\label{fig:MMM2}}
	\subfigure[27m\_vs\_30m \textbf{Super Hard}]{
		\includegraphics[width=0.325\linewidth]{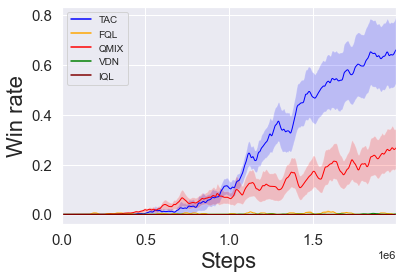}\label{fig:27m_vs_30m}}
% 	\vspace{-2mm}
	\subfigure[6h\_vs\_8z \textbf{Super Hard}]{
		\includegraphics[width=0.325\linewidth]{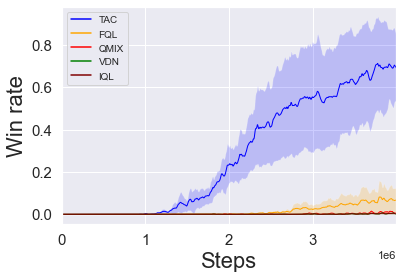}\label{fig:6h8z}}
	\subfigure[Corridor \textbf{Super Hard}]{
		\includegraphics[width=0.325\linewidth]{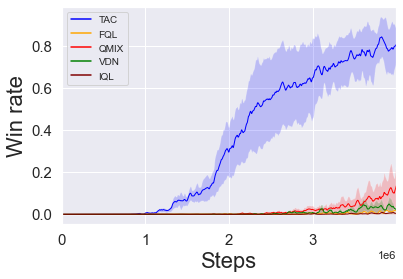}\label{fig:corridor}}
% 	\vspace{-3mm}
	\caption{Performance of different algorithms on different SMAC scenarios: \textcolor{blue}{TAC}, \textcolor{red}{QMIX}, \textcolor[rgb]{0,0.7,0}{VDN}, \textcolor{orange}{FQL}, \textcolor[rgb]{0.76, 0.13, 0.28}{IQL}. \label{fig:smac_exp}}
% 	\vspace{-3mm}
\end{figure*}

\paragraph{StarCraft II}
A challenging set of cooperative scenarios from the StarCraft  Multi-Agent Challenge (SMAC) \citep{samvelyan2019starcraft} is considered in this experiment. Scenarios in SMAC have been classified as \textbf{Easy, Hard and Super-hard} according to the performance of exiting algorithms on them. \textsc{Tesseract} (\textcolor{blue}{TAC} in plots) is compared with, \textcolor{red}{QMIX} \citep{rashid2018qmix}, \textcolor[rgb]{0,0.7,0}{VDN} \citep{sunehag_value-decomposition_2017}, \textcolor{orange}{FQL} \citep{chen2018factorized}, and \textcolor[rgb]{0.76, 0.13, 0.28}{IQL} \citep{tan_multi-agent_1993}, the latter being algorithms specifically designed for cooperative decentralised multi agent reinforcement learning. VDN and QMIX use monotonic approximations for learning the Q-function. FQL uses a pairwise factorized model to capture effects of agent interactions in joint Q-function, this is done by learning an inner product space for summarising agent trajectories. IQL ignores the multi-agentness of the problem and learns an independent per agent policy for the resulting non-stationary problem. \cref{fig:smac_exp} gives the win rate of the different algorithms averaged across five random runs. \cref{fig:2c_vs_64z} features 2c\_vs\_64zg, a hard scenario that contains two
allied agents but 64 enemy units (the largest in the SMAC domain) making the action space of
the agents much larger than in the other scenarios. \textsc{Tesseract} gains a huge lead over all the other algorithms in just one million steps. For the asymmetric scenario of 5m\_vs\_6m \cref{fig:5m_vs_6m}, \textsc{Tesseract}, QMIX, and VDN learn effective policies, similar behavior occurs in the heterogeneous scenarios of 3s5z \cref{fig:3s5z_smac} and MMM2\cref{fig:MMM2} with the exception of VDN for the latter. In 2s\_vs\_1sc in \cref{fig:2s_vs_1sc}, which requires a `kiting' strategy to defeat the spine crawler, \textsc{Tesseract} learns an optimal policy in just 100k steps. In the \textbf{super-hard} scenario of 27m\_vs\_30m \cref{fig:27m_vs_30m} having largest ally team of 27 marines, \textsc{Tesseract} again shows improved sample efficiency; this map also shows \textsc{Tesseract}'s ability to scale with the number of agents. Finally in the \textbf{super-hard} scenarios of 6 hydralisks vs 8 zealots \cref{fig:6h8z} and Corridor \cref{fig:corridor} which require careful exploration, \textsc{Tesseract} is the only algorithm which is able to find a good policy. It is observed that IQL doesn't perform well on any of the maps as it doesn't model agent interactions/non-stationarity explicitly. FQL loses performance possibly because modelling just pairwise interactions with a single dot product might not be expressive enough for joint-Q. Finally, VDN and QMIX are unable to perform well on many of the challenging scenarios possibly due to the monotonic approximation affecting the exploration adversely \citep{mahajan2019maven}.
% Additional plots and experiment details can be found in \cref{app:sc2} with \textbf{comparison with other baselines in \cref{app:additional_sc2}} including QPLEX\citep{wang2020qplex}, QTRAN\citep{ son2019qtran}, HQL\citep{matignon2007hysteretic}, COMA\citep{foerster2018counterfactual} . We detail the techniques used for stabilising the learning of tensor decomposed critic in \cref{app:techniques}.

\section{Related Work}
\label{sec:relw}
% \emph{Centralised Training with Decentralised Execution} (CTDE) paradigm allows MARL methods address computational tractability and communication constraints in multi-agent learning.
% Methods for single-agent RL, such as DQN and DDPG, have been extended to CTDE settings~\cite{tampuu_multiagent_2015,lowe2017multi}. 
% Recent works have developed both value-based methods~\cite{guestrin2002multiagent,mahajan2019maven} and policy gradient methods~\cite{foerster2018counterfactual,wei2018multiagent}. 
Previous methods for modelling multi-agent interactions include those that use coordination graph methods for learning a factored joint action-value estimation \cite{guestrin2002coordinated,guestrin2002context,bargiacchi2018learning}, however typically require knowledge of the underlying coordination graph. 
To handle the exponentially growing complexity of the joint action-value functions with the number of agents, a series of value-based methods have explored different forms of value function factorisation.
%VDN,  
VDN~\cite{sunehag_value-decomposition_2017} and QMIX~\cite{rashid2018qmix} use monotonic approximation with latter using a mixing network conditioned on global state. 
% QTRAN
QTRAN~\cite{son2019qtran} avoids the weight constraints imposed by QMIX by formulating multi-agent learning as an optimisation problem with linear constraints and relaxing it with L2 penalties. 
% MAVEN
MAVEN~\cite{mahajan2019maven} learns a diverse ensemble of monotonic approximations by conditioning agent $Q$-functions on a latent space which helps overcome the detrimental effects of QMIX’s monotonicity constraint on exploration. Similarly, Uneven~\cite{gupta2020uneven} uses universal successor features for efficient exploration in the joint action space. 
% QATTEN 
Qatten~\cite{Yang2020QattenAG} makes use of a multi-head attention mechanism to decompose $Q_{tot}$ into a linear combination of per-agent terms. RODE~\cite{wang2020rode} learns an action effect based role decomposition for sample efficient learning.
% Moreover, they do not offer any guarantees for convergence and optimality.
Policy gradient methods, on the other hand, often utilise the actor-critic framework to cope with decentralisation.
% MADDPG
MADDPG~\cite{lowe2017multi} trains a centralised critic for each agent. % that can be applied in mixed cooperative-competitive environments with continuous action spaces. 
% COMA
COMA~\cite{foerster2018counterfactual} %addresses the multi-agent credit assignment problem of the joint team reward by 
employs a centralised critic and a counterfactual advantage function.
These actor-critic methods, however, suffer from poor sample efficiency compared to value-based methods and often converge to sub-optimal local minima. While sample efficiency has been an important goal for single agent reinforcement learning methods ~\cite{mahajan2017symmetryde, mahajan2017symmetryl, kakade2003sample, lattimore2013sample}, in this work we shed light on attaining sample efficiency for cooperative multi-agent systems using low rank tensor approximation.

% \vspace{-2mm}
\emph{Tensor methods} have been used in machine learning, in the context of learning latent variable models~\cite{anandkumar2014tensor}, signal processing \cite{sidiropoulos2017tensor}, deep learning and computer vision~\cite{panagakis2021tensor}. 
% By going further than pairwise correlation, and forming the low-rank, observable moments of the data, it is possible to recover using tensor decomposition the parameters of a wealth of latent variable models such as hidden Markov models and latent Dirichlet allocation~\cite{anandkumar2014tensor}.
They provide powerful analytical tools that have been used for various applications, including the theoretical analysis of deep neural networks~\cite{cohen2016expressive}.
Model compression using tensors~\cite{cheng2017survey} has recently gained momentum owing to the large sizes of deep neural nets.
Using tensor decomposition within deep networks, it is possible to both compress and speed them up~\cite{cichocki2017tensor,t_net}. They allow generalization to higher orders~\cite{kossaifi2019efficient} and have also been used for multi-task learning and domain adaptation~\cite{bulat2019incremental}. In contrast to prior work on value function factorisation, \textsc{Tesseract} provides a natural spectrum for approximation of action-values based on the rank of approximation and provides theoretical guarantees derived from tensor analysis. Multi-view methods utilising tensor decomposition have previously been used in the context of partially observable single-agent RL~\cite{azizzadenesheli2016reinforcement,azizzadenesheli2019reinforcement}. There the goal is to efficiently infer the underlying MDP parameters for planning under rich observation settings~\cite{krishnamurthy2016pac}. Similarly \citep{bromuri2012tensor} use four dimensional factorization to generalise across Q-tables whereas here we use them for modelling interactions across multiple agents.
% \vspace{-2mm}	
\end{document}